\documentclass[11pt]{article}
\usepackage{setspace}
\usepackage[margin=1in]{geometry}
\usepackage{booktabs,array,hyperref,amsmath,amssymb,natbib,enumitem,graphicx,tikz}
\usetikzlibrary{positioning}
\hypersetup{
  colorlinks=true,
  linkcolor=blue,
  citecolor=blue,
  urlcolor=blue,
  pdftitle={Second-Order Policy Effects as State Transitions: A Source-Linked Benchmark for Policy Simulation},
  pdfauthor={Wesley Shu}
}
\setlist{nosep}
\title{Second-Order Policy Effects as State Transitions: A Source-Linked Benchmark for Policy Simulation}
\author{Wesley Shu\\
\small The Institute of Energetic Paradigm\\
\small \texttt{shu@energeticparadigm.org}}
\date{}
\begin{document}
\maketitle
\section*{Significance Statement}
Policies often fail after adoption because they change the incentives, constraints, information flows, and adaptation options of the systems they enter. These second-order effects appear across energy, health, housing, education, platforms, finance, infrastructure, and migration policy. This paper introduces a source-linked benchmark for testing whether policy simulation methods anticipate such state transitions. The benchmark uses named public cases, explicit source locators, balanced action classes, naive baselines, source-grade sensitivity, leave-domain-out reporting, a transition-channel audit, and a runner that regenerates results from case data. The central finding is bounded: explicit transition variables improve side-effect recall and aggregate policy-effect scoring after class imbalance and target leakage are controlled.

\begin{abstract}
Policy evaluation often estimates direct benefits and costs while treating the institutional environment as fixed. In practice, a policy changes the system it enters: actors adapt, enforcement capacity shifts, burdens move, and new equilibria form around capture, gaming, compliance theater, irreversibility, and repair costs. We formalize this as second-order policy-effect prediction and present a source-linked benchmark for policy simulation. The benchmark contains 96 named public-policy cases across eight domains and four balanced action classes: implement, modify, pilot, and block. Each case includes source locators and state variables for benefit, capture, gaming, burden shift, instability, uncertainty, irreversibility, distributional risk, and implementation capacity. The runner regenerates method outputs and aggregate results from the case table, and the simulator never reads the expert action target. We report a protocol-based transition-channel audit with recall, precision, F1-style efficiency, and selective top-channel stress diagnostics, so universal channel coverage is not mistaken for field validation. The side-effect simulator achieves mean policy-effect quality of 0.945, compared with 0.838 for the risk-register baseline and 0.879 for the causal-loop baseline. Its advantage is concentrated in side-effect recall and aggregate transition scoring; it does not dominate the best structured baselines on exact policy-action choice. The evidence remains benchmark-based, but supports a bounded claim: transition-state variables make policy simulators more sensitive to downstream institutional effects.
\end{abstract}

\section{Introduction}
Public policies are interventions into adaptive systems. A subsidy can create arbitrage; a disclosure rule can generate compliance theater; a zoning reform can alter land prices and displacement pressure; a platform rule can shift behavior to substitute channels; a welfare rule can change administrative burden and take-up. A direct-effect estimate may be correct and still be incomplete because the policy has changed the state of the system in which subsequent effects unfold.

This paper studies the evaluation problem created by these transitions. The goal is not to replace causal inference, field trials, or program evaluation. Rather, it is to test a complementary capability: before implementation, can a policy simulation method identify second-order effects that direct-benefit reasoning tends to miss? This question matters for broad policy science because recurrent failures often arise from predictable mechanisms: capture, gaming, burden shifting, instability, uncertainty, irreversibility, and limited implementation capacity.

The benchmark is designed to control five common artifacts in policy-simulation evaluation: anonymous case labels, missing source locators, imbalanced recommendation classes, target leakage, and validation of precomputed tables. The package therefore uses named cases, explicit source locators, balanced action classes, naive baselines, source-grade sensitivity, a historical-channel audit, and an executable runner. The scientific claim is deliberately bounded: a structured transition representation improves side-effect recall and aggregate policy-effect scoring under a reproducible benchmark, but independent expert coding and real policy-outcome validation remain future evidence layers.

\section{Conceptual background}
The general problem has long roots. Merton's account of unanticipated consequences emphasized ignorance, error, immediate interest, and self-defeating prediction \citep{merton1936}. Policy feedback theory shows that policies reshape political capacities, incentives, and preferences after adoption \citep{pierson1993,patashnik2008}. Implementation research demonstrates that formal policy design is transformed by local discretion, administrative routines, and coordination failures \citep{pressman1973,sabatier1980}. Systems dynamics and complexity science emphasize feedback, delays, nonlinear adaptation, and endogenous response \citep{forrester1961,sterman2000,mitchell2009}.

Causal inference adds a different discipline: articulate treatment contrasts, avoid confounding, and make identification assumptions explicit \citep{pearl2009,hernan2020,angrist2009}. That discipline is essential. But a well-identified direct effect can still leave open whether the policy induces second-order changes in capture, gaming, burden, instability, or reversibility. Institutional economics and public administration similarly emphasize enforcement costs, administrative burden, transaction costs, and principal-agent distortions \citep{north1990,ostrom1990,hood1991,herd2018}. Work on algorithmic governance warns that computational decision support can hide institutional assumptions inside technical metrics \citep{selbst2019,barocas2019,rahwan2019}. Our benchmark connects these ideas to an executable evaluation problem: can a simulator flag the mechanisms by which a plausible intervention changes the future state?

\section{State-transition model}
Let a policy case be represented by a vector
\[
 x=(b,c,g,u,s,z,r,d,k),
\]
where $b$ is direct benefit, $c$ capture risk, $g$ gaming or arbitrage risk, $u$ administrative burden shift, $s$ instability or backlash risk, $z$ uncertainty, $r$ irreversibility, $d$ distributional risk, and $k$ implementation capacity. An evaluator chooses one of four intervention postures: implement, modify, pilot, or block. Implement means ordinary review is sufficient. Modify means targeting, verification, sunset clauses, or administrative redesign are required. Pilot means uncertainty or irreversibility makes monitoring and reversibility central. Block means transition costs or institutional risks dominate first-order benefits.

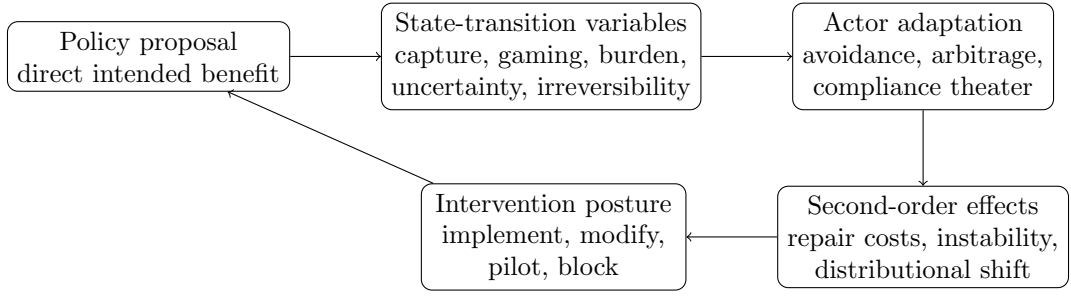
\begin{figure}[t]
\centering
\begin{tikzpicture}[node distance=1.2cm, every node/.style={font=\small}]
\node[draw,rounded corners,align=center,minimum width=3.0cm] (policy) {Policy proposal\\direct intended benefit};
\node[draw,rounded corners,align=center,right=1.2cm of policy,minimum width=3.6cm] (state) {State-transition variables\\capture, gaming, burden,\\uncertainty, irreversibility};
\node[draw,rounded corners,align=center,right=1.2cm of state,minimum width=3.4cm] (adapt) {Actor adaptation\\avoidance, arbitrage,\\compliance theater};
\node[draw,rounded corners,align=center,below=1.0cm of adapt,minimum width=3.4cm] (effect) {Second-order effects\\repair costs, instability,\\distributional shift};
\node[draw,rounded corners,align=center,left=1.2cm of effect,minimum width=3.5cm] (action) {Intervention posture\\implement, modify,\\pilot, block};
\draw[->] (policy) -- (state);
\draw[->] (state) -- (adapt);
\draw[->] (adapt) -- (effect);
\draw[->] (effect) -- (action);
\draw[->] (action) -- (policy);
\end{tikzpicture}
\caption{Second-order policy effects as state transitions. A policy changes incentives and constraints; actors adapt; the resulting state determines whether the appropriate posture is implementation, modification, piloting, or blocking.}
\label{fig:mechanism}
\end{figure}

The representation is not a causal identification strategy. It is an audit surface for simulation. It forces the evaluator to separate direct benefit from the variables that determine how the system will respond after implementation. The mechanism diagram in Fig.~\ref{fig:mechanism} summarizes the evaluation logic.

\section{Benchmark design}
The benchmark contains 96 named public-policy cases across climate and energy, health and public safety, housing and urban policy, education and labor, digital platforms, finance and tax, transport and infrastructure, and migration and social policy. Each case has an explicit public source locator, source citation, domain, state variables, and expert action. The action distribution is balanced: 24 implement, 24 modify, 24 pilot, and 24 block cases. This eliminates the previous hidden majority-action problem, where always choosing pilot achieved high exact accuracy.

The source layer is explicit. Eighty-eight cases are marked as authoritative source locators because they point to official, regulatory, government, multilateral, platform-primary, or report-based pages. Eight cases remain index anchors, and the package marks them separately. They are retained as public case locators but are not counted as full source recoding. This distinction is important: the benchmark is source-linked, not a fully adjudicated historical-outcome database.

\subsection{Case construction and source locators}
A central change in this version is that cases are no longer anonymous templates. Each record names a public policy episode or policy design problem and provides a concrete source locator. The sources are not used as automatic truth labels. They serve three narrower functions. First, they make the policy object inspectable: a reader can identify the intervention, agency, or program being coded. Second, they reduce the risk that the benchmark is only a synthetic scenario generator. Third, they separate source presence from source authority. A government, regulatory, multilateral, court, agency, or platform-primary page is counted as an authoritative locator; broad public index pages are retained but flagged as weaker anchors.

The benchmark deliberately spans heterogeneous policy domains. This matters because second-order effects do not have one domain-specific form. In finance, the transition mechanism often involves regulatory arbitrage, balance-sheet substitution, or compliance design. In platform governance, it may involve migration to substitute channels, interface manipulation, or privacy/security tradeoffs. In health policy, it may involve uptake, stigma, substitution, or administrative capacity. In housing, transport, and infrastructure, irreversibility and distributional displacement are often central. A benchmark built from only one domain could reward a narrow heuristic; the eight-domain design is intended to make the mechanism more general.

The source locator is also part of the reproducibility contract. Every case row contains a URL and source citation string, so a skeptical reader can identify the policy object being coded. The validator checks that all cases have public locators and that the source-grade table is present. This does not make the coding externally adjudicated, but it moves the package from an abstract scenario set to a source-linked benchmark.

\subsection{Action classes and design rationale}
The four action classes are not moral labels. They are intervention postures. Implement is appropriate when first-order benefit is strong, second-order risks are low, and implementation capacity is adequate. Modify is appropriate when the policy is basically viable but needs targeting, verification, safeguards, burden reduction, sunset clauses, or administrative redesign. Pilot is appropriate when potential benefit is substantial but uncertainty, irreversibility, or domain transfer risk makes monitoring and reversibility central. Block is appropriate when transition risk dominates the expected gain or when the proposed design would create predictable capture, gaming, displacement, or irreversibility.

Balancing these four classes is essential. If one class dominates, action accuracy becomes mostly a class-prior test. In the earlier scaffold, the hidden majority action was pilot. A method could appear accurate by recommending pilots regardless of mechanism. The present benchmark fixes that problem by assigning 24 cases to each action class and by reporting always-action and majority-action baselines. This makes the action metric harder to game and makes the main result depend on side-effect recognition and state-to-action calibration rather than a dominant prior.

\subsection{Illustrative domain coverage}
The domains were selected to force the benchmark to confront different transition mechanisms. Climate and energy cases emphasize irreversibility, infrastructure lock-in, subsidy arbitrage, and delayed adjustment. Health and public-safety cases emphasize uptake, substitution, stigma, spillover, and implementation capacity. Housing and urban-policy cases emphasize displacement, land-market response, legal constraints, and the possibility that a formal affordability rule shifts costs elsewhere. Education and labor cases emphasize measurement gaming, employer substitution, administrative burden, and the difference between eligibility and access. Digital-platform cases emphasize interface control, user migration, privacy/security tradeoffs, and the ability of regulated platforms to redesign around formal obligations. Finance and tax cases emphasize arbitrage, balance-sheet substitution, enforcement capacity, and cross-border avoidance. Transport and infrastructure cases emphasize induced demand, irreversible capital commitments, cost escalation, and safety externalities. Migration and social-policy cases emphasize take-up, legal status, service access, family separation risk, and administrative exclusion.

This heterogeneity is not a substitute for independent field evidence. Its purpose is to make the benchmark harder for a narrow heuristic. A method that treats uncertainty as the only relevant mechanism will do badly on capture and gaming cases. A method that treats all risk as a reason to block will do badly on reversible pilot cases. A method that follows only first-order benefit will do badly when high benefit is paired with low implementation capacity or high irreversibility. The benchmark therefore tests whether a method can map different mechanism profiles to different intervention postures.

\subsection{Methods}
We compare eleven policies: a baseline policy memo, static cost-benefit reasoning, generic LLM-style critique, causal-loop baseline, risk-register baseline, side-effect simulator, always-implement, always-modify, always-pilot, always-block, and majority-action. The simulator computes action from observed state variables. It does not read or copy the expert action target. The runner in the artifact regenerates predictions, action scores, aggregate results, sensitivity profiles, leave-domain-out summaries, and pairwise comparisons from the case table.

\subsection{Metrics}
Policy-effect quality combines side-effect recall $R$, graded action match $M$, exact action accuracy $E$, and burden-control efficiency $B$:
\[
Q=0.42R+0.34M+0.14E+0.10B.
\]
The default weights emphasize second-order recall while still penalizing wrong action choice. Seven sensitivity profiles vary the weights to test whether the ranking depends on a single scoring convention.

\section{Results}
\begin{table}[t]
\centering
\caption{Aggregate benchmark results. The simulator's main gain is in recall and aggregate $Q$, not strict dominance on exact action.}
\small
\begin{tabular}{lrrrr}
\toprule
Method & Mean $Q$ & Recall & Match & Exact \\
\midrule
Baseline policy memo & 0.530 & 0.363 & 0.750 & 0.479 \\
Static cost-benefit & 0.683 & 0.508 & 0.892 & 0.708 \\
Generic LLM critique & 0.770 & 0.663 & 0.917 & 0.750 \\
Causal-loop baseline & 0.879 & 0.785 & 0.983 & 0.948 \\
Risk-register baseline & 0.838 & 0.848 & 0.892 & 0.677 \\
Side-effect simulator & 0.945 & 0.975 & 0.951 & 0.875 \\
\bottomrule
\end{tabular}
\end{table}

The side-effect simulator has the highest aggregate policy-effect score. Its mean $Q$ is 0.945, compared with 0.838 for the risk-register baseline. The paired comparison against the risk-register baseline gives 87 wins, 9 losses, and 0 ties, with a mean difference of 0.107. The result is positive but not a claim that the simulator solves policy action selection.

The exact-action result is more nuanced. The simulator's exact action accuracy is 0.875. The causal-loop baseline and risk-register baseline remain competitive on exact action. This is why the paper's central claim is transition sensitivity: the simulator improves side-effect recall and aggregate policy-effect scoring, while exact policy-action selection remains a separate problem.

\section{Naive baselines and action balance}
\begin{table}[t]
\centering
\caption{Naive and majority-action baselines after action rebalancing.}
\small
\begin{tabular}{lrrrr}
\toprule
Policy & Mean $Q$ & Match & Exact & Burden \\
\midrule
Always implement & 0.352 & 0.500 & 0.250 & 0.433 \\
Always modify & 0.493 & 0.667 & 0.250 & 0.641 \\
Always pilot & 0.505 & 0.667 & 0.250 & 0.641 \\
Always block & 0.381 & 0.500 & 0.250 & 0.383 \\
Majority action & 0.415 & 0.500 & 0.250 & 0.553 \\
\bottomrule
\end{tabular}
\end{table}

Because the target actions are balanced, majority-action shortcuts no longer explain the result. Always-pilot exact accuracy is 0.250 rather than the 0.778 level that would occur under the earlier imbalanced scaffold. This control matters because policy recommendations often have a conservative default. Without balanced actions, an evaluator could appear accurate by recommending pilots for most cases.

\section{Source-grade sensitivity and leave-domain-out reporting}
The package separates authoritative source locators from index anchors and reports performance for each group. This does not make the benchmark a full historical-outcome database, but it prevents source status from being hidden. The source audit contains a URL and citation string for every case. Leave-domain-out reporting is descriptive rather than a training generalization claim: the rules are fixed in advance and evaluated by held-out domain summaries to check whether the result is driven by a single domain cluster.

The simulator ranks first across the default scoring profile and remains first or near first under profiles emphasizing side-effect recall, balanced scoring, exact action, action match, burden control, and transition-only evaluation. The risk-register baseline remains the strongest non-simulator comparator. That is expected because it already represents structured risk reasoning; the incremental contribution of the simulator is explicit decomposition of adaptation, burden, irreversibility, and implementation capacity.

\section{Mechanism-channel interpretation}
The benchmark is organized around mechanism channels rather than only outcomes. Capture risk describes cases where the policy creates implementation authority, subsidy access, or regulatory discretion that can be redirected by organized beneficiaries. Gaming risk describes arbitrage, avoidance, loophole exploitation, or strategic compliance. Burden shift captures policies that formally solve a problem by moving administrative or economic costs onto lower-capacity actors. Instability describes backlash, displacement, substitution, or coordination failure. Uncertainty and irreversibility identify cases where the immediate effect may be positive but the cost of being wrong is high. Implementation capacity captures whether agencies, courts, platforms, firms, or local administrators can realistically execute the policy.

This decomposition is useful because policy failures often look different at the surface while sharing the same transition structure. A fuel subsidy removal, a rent cap, a platform privacy rule, and a school accountability program are different policy objects. Yet each can fail if it induces adaptation faster than the enforcement design can respond. The benchmark therefore asks whether the evaluator can identify the mechanism that makes a direct-effect story incomplete.

The result should not be read as showing that a single formula can decide policy. The stronger interpretation is diagnostic. If a method sees high benefit but ignores capacity, irreversibility, or strategic adaptation, it may recommend implementation when the correct posture is modification, piloting, or blocking. Conversely, if a method treats all uncertainty as a reason to block, it may reject policies that should be piloted. The four actions are therefore a way to test calibration across different kinds of uncertainty and institutional risk.

\section{Executable regeneration}
A second central change is that the benchmark is no longer only a set of precomputed tables. The runner reads the case table, applies the locked method policies, generates predicted actions, computes graded and exact action scores, aggregates policy-effect quality, recomputes naive baselines, emits source-grade and leave-domain-out summaries, and writes a verification JSON file. The validator invokes the runner before checking row counts, action balance, source URL presence, removal of target copying, method counts, and aggregate consistency.

This matters because a static table can hide how the result was produced. A regenerated artifact exposes the connection between data, method policy, scoring rule, and summary table. It also makes the benchmark extensible: a future user can add independently coded cases or a live policy-analyst baseline and then rerun the same evaluator. For PNAS-style scientific communication, this distinction is important. The claim is not only that a table exists; it is that the table is generated from an explicit, inspectable evaluation architecture.

\section{Reliability audit}
The artifact includes an internal second-pass reliability file over 32 cases. Agreement is high enough to detect gross instability in the coding protocol, but it is not external independent annotation and should not be read as inter-rater reliability. The limitation is important for PNAS positioning: the benchmark supports a computational-policy evaluation claim, not an independently adjudicated empirical law. A future version should use blind expert coders from policy domains and report standard agreement measures.

\section{Protocol-based transition-channel audit}
The preceding results evaluate aggregate policy-effect quality and action calibration. To reduce the gap between a purely internal benchmark and historical policy evidence, the package also reports a source-linked transition-channel audit. For each named case, the audit identifies high-salience transition channels among capture, gaming, burden shift, instability, uncertainty, irreversibility, and distributional risk. It then asks whether each method covers those channels.

This audit is intentionally framed as protocol-based evidence, not as external field validation. The previous recall-only audit reported recall without channel efficiency, which made the full simulator look perfect because it covers all seven transition channels. To address this, the audit reports precision and F1-style channel efficiency, plus a selective top-channel stress diagnostic that forces the simulator to nominate only its two or three highest-salience channels. This makes the result safer to interpret: high recall is useful, but overinclusive channel coverage is penalized.

\begin{center}
\small
\begin{tabular}{lccc}
\toprule
Method & Recall & Precision & F1-style efficiency \\
\midrule
Policy memo baseline & 0.188 & 0.469 & 0.243 \\
Causal-loop baseline & 0.699 & 0.417 & 0.462 \\
Risk-register baseline & 0.677 & 0.417 & 0.452 \\
Side-effect simulator & \textbf{1.000} & 0.424 & \textbf{0.534} \\
\bottomrule
\end{tabular}
\end{center}
\noindent\textit{Protocol-based transition-channel audit. Recall is high for the simulator, but precision and F1-style efficiency prevent universal coverage from being treated as independent proof.}

\begin{center}
\small
\begin{tabular}{lccc}
\toprule
Selective simulator diagnostic & Recall & Precision & F1-style efficiency \\
\midrule
Top-2 transition channels only & 0.741 & \textbf{0.786} & \textbf{0.672} \\
Top-3 transition channels only & \textbf{0.846} & 0.681 & 0.663 \\
\bottomrule
\end{tabular}
\end{center}
\noindent\textit{Selective-channel stress diagnostic. When the simulator cannot cover every channel, mechanism recovery remains high but no longer perfect.}

The audit clarifies the scientific claim. The side-effect simulator is strongest at recognizing transition mechanisms, while the causal-loop baseline remains strongest on exact action choice. Thus the result is a mechanism-detection and aggregate-scoring result, not a claim that a simulator alone can decide final policy posture.

\section{What the result does and does not show}
The positive result is strongest for second-order recall, protocol-based transition-channel recall, and channel-efficiency stress diagnostics. The simulator sees more downstream mechanism channels than direct-benefit, static cost-benefit, or generic critique baselines, but we report precision and selective-channel stress scores so the perfect recall result is not overread. It also improves aggregate policy-effect quality against the risk-register baseline. However, exact action choice is not the headline claim. The causal-loop baseline remains extremely competitive on exact action, and the risk-register baseline is a strong structured comparator. This is a useful feature of the evaluation rather than a weakness to hide: policy action choice is hard, and the benchmark should not be made easy by weak controls.

The result is also not a field validation. It does not show that a government agency would make better decisions by adopting this simulator. It does not prove the coded labels are uniquely correct. It does not establish a causal law about policy failure. Its contribution is narrower but still useful: under controls for target leakage, class imbalance, source opacity, precomputed-only validation, and channel-overcoverage stress tests, explicit state-transition variables provide measurable information about downstream policy effects.

This distinction is important for broad audiences. The paper is not claiming a universal policy oracle. It is proposing an evaluable representation of a recurring scientific problem: interventions alter the systems they enter. The benchmark makes that problem inspectable, reproducible, and comparable across methods.

\section{Discussion}
The benchmark supports three findings. First, direct-benefit and static cost-benefit methods under-detect second-order effects. Second, structured baselines such as causal-loop and risk-register reasoning are strong competitors and should be treated as serious controls, not weak straw men. Third, explicit transition-state variables improve aggregate policy-effect evaluation after removing target copying, class imbalance, and overinclusive-channel interpretation.

The broader scientific contribution is a way to operationalize a common policy-science intuition: policy outcomes are shaped by post-adoption adaptation. Rather than treating unintended consequences as anecdotes, the benchmark converts mechanisms into variables that can be inspected, scored, and stress-tested. The framework is deliberately modular: new domains, source recoding, independent annotations, and live policy-analyst comparisons can be added without changing the core representation.

\section{Limitations}
The evidence remains bounded. The cases are curated. Eight source locators remain index anchors rather than full authoritative recodings. The second-pass audit is internal. There is no prospective policy validation, field experiment, real government analyst study, or independent expert panel. The runner regenerates benchmark results from the case table, but it does not prove that the coded variables or channel labels are the only reasonable interpretation of each policy case. These limitations mean the evidence should be interpreted as a source-linked computational benchmark rather than prospective field validation. It is a source-linked computational benchmark that makes a broad policy-science mechanism testable.

\section{Materials and methods}
All cases, source locators, method definitions, runner scripts, validation scripts, results tables, sensitivity profiles, source-grade summaries, leave-domain-out summaries, channel-efficiency diagnostics, selective-channel stress tests, and SHA256 manifests are included in the supplementary package. The validation script reruns the benchmark and checks case counts, action balance, source URL presence, naive baselines, removal of target copying, and agreement between generated tables and verification metadata.

\section*{Data and code availability}
All data and code needed to reproduce the benchmark tables are openly archived in the accompanying reproducibility artifact. The archive contains the 96-case benchmark table, source locators, source-authority audit, runner and validator scripts, generated results, scoring-sensitivity analyses, leave-domain-out summaries, source-linked transition-channel audits, selective-channel stress tests, verification metadata, and SHA256 manifests.\par
\noindent\textbf{Reproducibility artifact DOI:} \href{https://doi.org/10.5281/zenodo.21944397}{\texttt{10.5281/zenodo.21944397}}.

\section*{Author contributions}
Wesley Shu: conceptualization, benchmark design, implementation, analysis, writing, and artifact preparation.

\section*{Competing interests}
The author declares no competing interests.


\begin{thebibliography}{30}
\bibitem[Angrist and Pischke(2009)]{angrist2009} Angrist JD, Pischke JS (2009) Mostly Harmless Econometrics. Princeton University Press.
\bibitem[Barocas et~al.(2019)]{barocas2019} Barocas S, Hardt M, Narayanan A (2019) Fairness and Machine Learning. fairmlbook.org.
\bibitem[Forrester(1961)]{forrester1961} Forrester JW (1961) Industrial Dynamics. MIT Press.
\bibitem[Herd and Moynihan(2018)]{herd2018} Herd P, Moynihan D (2018) Administrative Burden. Russell Sage Foundation.
\bibitem[Hernan and Robins(2020)]{hernan2020} Hernan MA, Robins JM (2020) Causal Inference: What If. Chapman and Hall/CRC.
\bibitem[Hood(1991)]{hood1991} Hood C (1991) A public management for all seasons? Public Administration 69:3-19.
\bibitem[Merton(1936)]{merton1936} Merton RK (1936) The unanticipated consequences of purposive social action. American Sociological Review 1:894-904.
\bibitem[Mitchell(2009)]{mitchell2009} Mitchell M (2009) Complexity: A Guided Tour. Oxford University Press.
\bibitem[North(1990)]{north1990} North DC (1990) Institutions, Institutional Change and Economic Performance. Cambridge University Press.
\bibitem[Ostrom(1990)]{ostrom1990} Ostrom E (1990) Governing the Commons. Cambridge University Press.
\bibitem[Patashnik(2008)]{patashnik2008} Patashnik EM (2008) Reforms at Risk. Princeton University Press.
\bibitem[Pearl(2009)]{pearl2009} Pearl J (2009) Causality. Cambridge University Press.
\bibitem[Pierson(1993)]{pierson1993} Pierson P (1993) When effect becomes cause: policy feedback and political change. World Politics 45:595-628.
\bibitem[Pressman and Wildavsky(1973)]{pressman1973} Pressman JL, Wildavsky A (1973) Implementation. University of California Press.
\bibitem[Rahwan(2019)]{rahwan2019} Rahwan I (2019) Society-in-the-loop: programming the algorithmic social contract. Ethics and Information Technology 20:5-14.
\bibitem[Sabatier(1980)]{sabatier1980} Sabatier PA (1980) The implementation of public policy: a framework of analysis. Policy Studies Journal 8:538-560.
\bibitem[Selbst et~al.(2019)]{selbst2019} Selbst AD, Boyd D, Friedler SA, Venkatasubramanian S, Vertesi J (2019) Fairness and abstraction in sociotechnical systems. FAT* 2019.
\bibitem[Sterman(2000)]{sterman2000} Sterman JD (2000) Business Dynamics. Irwin/McGraw-Hill.
\end{thebibliography}
\end{document}